\documentclass{article}

\usepackage{PRIMEarxiv}

\usepackage[utf8]{inputenc} 
\usepackage[T1]{fontenc}    
\usepackage{hyperref}       
\usepackage{url}            
\usepackage{booktabs}       
\usepackage{amsmath,amssymb,amsfonts,amstext}
\usepackage{nicefrac}       
\usepackage{microtype}      
\usepackage{lipsum}
\usepackage{fancyhdr}       
\usepackage{graphicx}       
\usepackage{wrapfig}
\usepackage{cleveref}
\usepackage{etoolbox}
\usepackage{makecell}

\title{No Captions, No Problem: \\
Captionless 3D-CLIP Alignment with Hard Negatives via CLIP Knowledge and LLMs
}

\author{
  Cristian Sbrolli, Matteo Matteucci \\
  Department of Electronics Information and Bioengineering \\
  Politecnico di Milano, Italy \\
  \texttt{\{name.surname\}@.polimi.it} \\
}

\begin{document}
\maketitle

\begin{abstract}
In this study, we explore an alternative approach to enhance contrastive text-image-3D alignment in the absence of textual descriptions for 3D objects. We introduce two unsupervised methods, $I2I$ and $(I2L)^2$, which leverage CLIP knowledge about textual and 2D data to compute the neural perceived similarity between two 3D samples. We employ the proposed methods to mine 3D hard negatives, establishing a multimodal contrastive pipeline with hard negative weighting via a custom loss function. We train on different configurations of the proposed hard negative mining approach, and we evaluate the accuracy of our models in 3D classification and on the cross-modal retrieval benchmark, testing image-to-shape and shape-to-image retrieval. Results demonstrate that our approach, even without explicit text alignment, achieves comparable or superior performance on zero-shot and standard 3D classification, while significantly improving both image-to-shape and shape-to-image retrieval compared to previous methods.
\end{abstract}


\section{Introduction}
In recent years, multimodal feature alignment gathered increasing interest in the scientific community. This was primarily sparked by the introduction of CLIP (Contrastive Language-Image Pretraining)~\cite{clip}, which successfully aligns images and text through contrastive multimodal learning, obtaining impressive results in many downstream tasks. 

Extending CLIP to 3D data is however a challenging task, primarily due to the scarcity of well-annotated text-2D-3D triplets required. Indeed, although image-3D datasets are relatively easy to obtain, for instance through rendering, there is a lack of data with well-curated and structurally meaningful descriptions for 3D models.

Initial efforts to address this issue aligned images and 3D objects, either without text or with template text prompts filled with category names or metadata-derived descriptions. More recent works have leveraged Large Multimodal Models (LMMs) capable of processing images to generate image captions, thereby generating text descriptions for the 3D object.  However, these generated texts often lack descriptiveness and discriminativeness, and suffer from inherent LMM biases like counting inability and a propensity to fill in missing information. They also encounter problems associated with image-based queries, including the incorporation of information not obtainable from the point cloud model, such as colours, textures, and materials in the 3D model.

In light of the limitations associated with LMM-generated prompts, this study explores an alternative approach to enhance contrastive text-2D-3D alignment in the absence of labelled textual descriptions for 3D objects. Specifically, we propose to leverage CLIP knowledge about textual and 2D data to compute similarities between point cloud models, then exploit those similarities to mine hard negative samples, enhancing the contrastive training. We introduce two unsupervised approaches for computing the neural-perceived similarity between 3D samples, $I2I$ and $(I2L)^2$. The first computes 3D similarities by only leveraging object views and the CLIP image encoder. The second method we propose leverages instead both images and LLM-generated texts, while addressing both the limitations of image-based approaches and the low-descriptiveness of image-captioning generated prompts. To this end, we transpose the 3D shape similarity problem into an image-text similarity space, where each object view is characterized by its similarity vector in relation to a set of fixed and detail-rich LMM-generated category prompts that do not refer to color, texture, or material. 

We establish a text-image-3D contrastive pipeline that uses precomputed 3D similarities to perform hard negative sampling, and we instantiate it in different configurations by employing the proposed similarities. We rigorously test the obtained models following evaluation benchmarks and community protocols for text-image-3D contrastive learning on multiple datasets, both real and synthetic. We assess the macro-structure of the latent space by testing on 3D classification (both zero-shot and fine-tuned). We evaluate instead the fine-grained understanding of intra-category details and the alignment of 3D and images by a cross-modal retrieval benchmark, testing image-to-shape and shape-to-image retrieval. Our experiments demonstrate that our approach achieves superior or comparable performance on 3D classification, while obtaining a significant improvement in both image-to-shape and shape-to-image retrieval compared to state-of-the-art methods.
\begin{figure*}
\begin{center}
\includegraphics[width=0.97\linewidth]{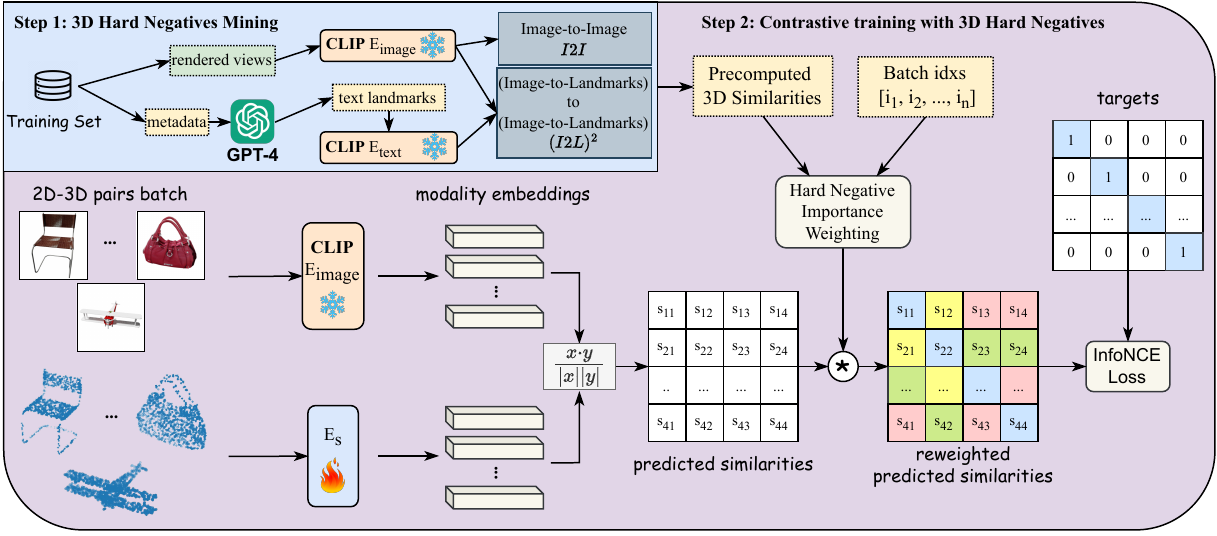}
\end{center}
   \caption{Overview of our proposed approach. We first precompute 3D samples similarities through our proposed neural similarity metrics, then we use the obtained scores to enhance the contrastive training with hard negatives.}
\label{fig:short}
\end{figure*}

\section{Related Works}
\label{sec:relworks}
Over the past decade, Contrastive Representation Learning (CRL)~\cite{contrastivelearning} has emerged as a robust self-supervised learning framework. It builds a well-structured representation space by learning to discern similar and dissimilar samples. The advent of InfoNCE loss~\cite{infonce1,infonce2} and its application in CLIP (Contrastive Language Image Pretraining)~\cite{clip} has further propelled CRL’s popularity. CLIP, aligning images and text in the latent space of two separate modality encoders, has shown remarkable results in various downstream tasks, making it a key component in models like LLaVA~\cite{llava} and DALLE-2~\cite{dalle2}. Subsequent research has enhanced the CLIP methodology with soft labels~\cite{softclip}, sigmoid loss~\cite{sigclip}, data filtering, and hard negatives~\cite{filterdisthard}, leading to improved text-image alignment.

Building on the success of CLIP with text-image data, researchers have expanded this approach to other modalities, including audio~\cite{audioclip}, video~\cite{videoclip}, and even imu, depth, and heat maps~\cite{imagebind}. The extension to 3D arrived with initial works obtaining the alignment of voxels and images~\cite{ic3d}, followed by more efficient 3D representations. Early works aligning point clouds with other modalities, such as PointCLIP~\cite{pointclip} and its successor, PointCLIP v2~\cite{pointclipv2}, use the pretrained vision encoder from CLIP to process point clouds by projecting them onto multi-view depth maps. CLIP2Point~\cite{clip2point} also trains on depth maps projected from point clouds, but introduces a fully trainable depth encoder and an intra-modality loss for aligning different depth maps of the same 3D sample, thereby enhancing invariance to view changes. ULIP~\cite{ulip} presents and successfully trains a pure text-image-3D point cloud model, introducing a point cloud encoder instead of using projected depth maps, and outperforms previous works in all benchmarks. After ULIP, as with CLIP, research in this field has begun to explore ways to improve the model’s ability to align concepts from 3D with other modalities. ReCON~\cite{recon} achieves this by combining a generative paradigm along with the contrastive loss to ensure the latent space retains more information about the 3D shape. While effective, this approach is less parameter-efficient, as it employs an additional generative branch and module. ULIP 2~\cite{ulip2} maintains efficiency by focusing instead on generating more informative text captions. Indeed, while ULIP used metadata and class categories to create text labels for image-text-3D triplets, ULIP 2 employs BLIP-2~\cite{blip2} to pre-generate per-sample text captions for 2D views of the training set objects. This allows for a more informative contrastive training between text and shape embeddings. However, the use of a general-purpose LMM means that the captions inherit both the strengths and limitations of the model. For instance, by analyzing the generated texts (which are publicly available), it is evident that the model lacks the specialization required to describe relevant structural elements of the considered objects. Furthermore, the captions often describe the object in a generic manner, sometimes hallucinating scenes biased toward the object’s function or background color. Another challenge is the reference to color, textures, and materials in the captions, which are not visible from the point cloud. We further discuss and exemplify these aspects of the ULIP\nobreakdash-2 captions in the supplementary.

In light of the aforementioned issues with previous models, the fundamental problem for this work is to remain parameter-efficient by not adding more models/modules, while still incorporating additional knowledge to achieve a more informative 3D contrastive training.

\section{Method}
\label{sec:method}
\subsection{Preliminaries: Contrastive Training and Hard Negatives}
The objective of contrastive multimodal training, as proposed in CLIP~\cite{clip}, is to accurately match the same sample from different modalities among all the samples in a batch. Formally, the model takes a batch of N coupled samples $(x_{m_1}, x_{m_2})$ from different modalities $m_1$ and $m_2$ e.g. couples of images and their corresponding captions. Two separate modality encoders $E_{m_1}$ and $E_{m_2}$ process the batch producing L2-normalized modality embeddings for the batch, $e_{m_1}$ and $e_{m_2}$, each with shape $(N, F)$ with $F$ being the number of latent features.
The objective of the model is that of matching, through cosine similarity, corresponding samples from different modalities, thus aligning the corresponding embeddings in the latent space. The contrastive self-supervised loss is then defined as:
\begin{equation}
\operatorname{\mathcal{L}_{(m_1,m_2)}} = \frac{1}{2} \operatorname{\mathcal{L}_{m_1 \to m_2}} + \frac{1}{2} \operatorname{\mathcal{L}_{m_2 \to m_1}},\quad \operatorname{\mathcal{L}_{m_i \to m_j}} = - \frac{1}{N}  \sum\limits_{k=1}^{N} \log{\frac{\exp{\left(\nicefrac{e_{m_i}^k \cdot e_{m_j}^k}{\tau}\right)}} {\sum\limits_{s=1}^{N}\exp{\left(\nicefrac{e_{m_i}^k \cdot e_{m_j}^s}{\tau}\right)}}},
\label{eq:contrlossm1m2}
\end{equation}
with $\cdot$ being the inner product, $\tau$ a learnable temperature parameter stored as a log-parameterized scalar, and the apices indicize elements in the batch. 

If all samples in the batch are vastly different, it becomes easier for the model to match samples, resulting in the model learning less detailed features. Conversely, if the batch contains similar samples, the model faces a challenge in discriminating the correct one, thereby forcing it to learn more detailed features. These samples are named as hard negatives, and their mining and sampling is a well-studied topic in classic representation learning~\cite{hardnegrepreslearn1,hardnegrepreslearn2}. More recently, it has been applied in the context of contrastive learning, proving successful in aligning 2D images~\cite{hardnegsampling} and text-2D~\cite{filterdisthard}.  Yet, it remains unexplored in text-2D-3D contrastive representation learning, where the first step is to address hard negative mining, a non-trivial task with 3D data. In the following, we approach this problem by defining neural similarities to extract hard samples for a given 3D object in a dataset.

\subsection{Similarities for 3D Hard Negative Mining: $I2I$, $(I2L)^2$}
\begin{figure*}
\begin{center}
\includegraphics[width=\linewidth]{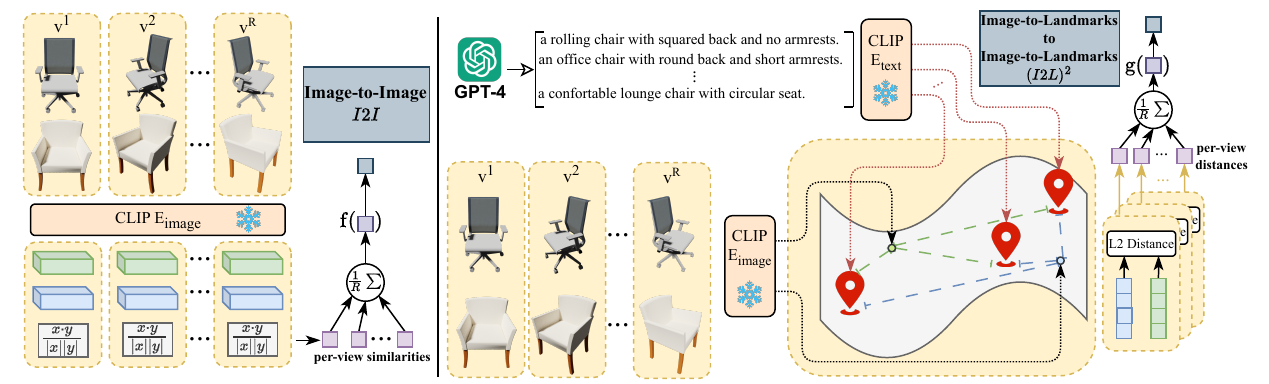}
\end{center}
   \caption{Our proposed similarity metrics for 3D hard negative mining.}
\label{fig:similarities}
\end{figure*}

In preliminary experiments, reported in the supplementary, we tested with established 3D similarity metrics like Chamfer Distance (CD) and Earth Mover’s Distance (EMD). However, these metrics did not yield satisfactory outcomes.
While they excel at overall correspondence assessment, they tend to overlook finer details, favoring denser regions~\cite{emdcmdproblems}. Moreover, they are not context-aware, thus they are not able to extract and compare context-relevant features which are paramount for the extraction of semantically challenging negatives. These limitations mainly impact the matching of local geometric structures, such as small or thin elements, holes and curvatures, all of which are crucial features in describing and distinguishing 3D samples. We propose then to build neural similarities capable of capturing relevant context-aware features by exploiting the semantically rich CLIP latent space. We introduce two similarity measures, depicted in \Cref{fig:similarities}, built upon the publicly available  OpenCLIP~\cite{openclip} model in its ViT\nobreakdash-B\nobreakdash-32 version.

\begin{figure*}
\begin{center}
\includegraphics[width=\linewidth]{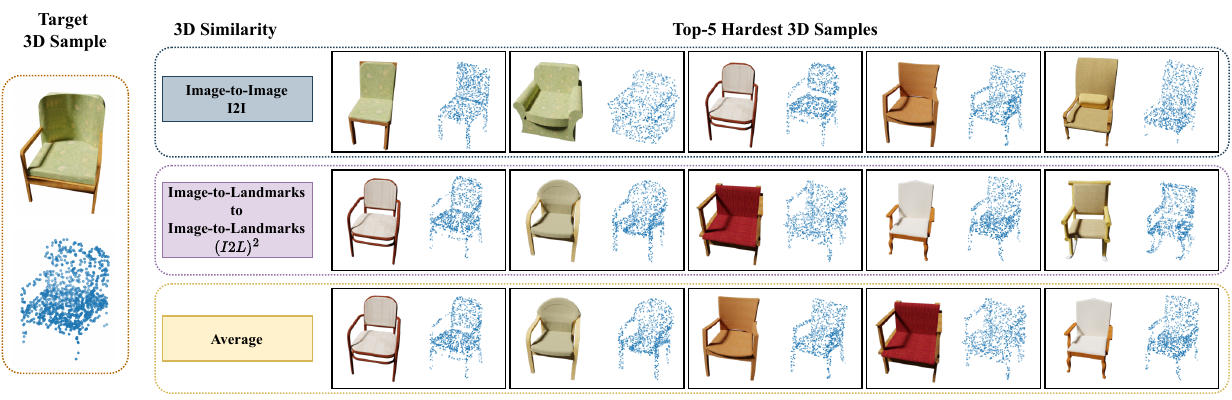}
\end{center}
   \caption{3D-to-3D retrieval using our similarities on a chair sample from ShapeNet dataset.}
\label{fig:3Dto3DRetrieval}
\end{figure*}

\noindent\textbf{ $\mathbf{I2I}$ - Image to Image 3D similarity:}
Our first proposed strategy primarily utilizes images, based on the assumption that 2D views provide sufficient information for the 3D object. For each 3D sample $x_{3D}^i$, we consider its set of R rendered views $\{v^{i,r}\}_{r=0}^R$ from a set of fixed camera poses for all samples. We employ the pretrained OpenCLIP image encoder $E_i$ to compute the image embedding $e_{image}^{k,r}$ for each view. With a sufficient number of views, these embeddings serve as neural descriptors of the 3D object, capturing features from all sides. Given these descriptors for any pair of point cloud samples $(x_{3D}^1,x_{3D}^2)$, we compute the $I2I$ 3D similarity as:
\begin{equation}
\operatorname{3D{\text -}Sim_{I2I}(x_{3D}^1,x_{3D}^2)} = f\left(\frac{1}{V} \sum\limits_{r=1}^V{e_{image}^{1,r} \cdot e_{image}^{2,r}}\right),\quad f(x) = \frac{(x+1)}{2}
\end{equation}
which, considering L2-normalized embeddings, is the mean cosine similarity between corresponding views of the two objects, shifted to $[0,1]$ with $f(x)$. As demonstrated in \Cref{fig:3Dto3DRetrieval}, this method effectively matches similar items by leveraging the fact that while CLIP is trained to align images and text, it also implicitly learns to align images containing similar concepts and features. However, in \Cref{fig:3Dto3DRetrieval} we also show that this approach still encounters the same color-texture-material issue discussed in \Cref{sec:relworks}, where objects with similar texture result more similar than structurally similar samples.

\textbf{ $\mathbf{(I2L)^2}$ - (Image to Landmarks) to (Image to Landmarks) 3D similarity:} We propose a second strategy with two aims: (1) solving the colour-texture-material limitation and (2) leveraging both the image and text knowledge of CLIP derived from its large scale training. First, we need to define how to efficiently use text to describe 3D objects, so that we can leverage CLIP text encoder to our advantage. However, as mentioned in \Cref{sec:relworks}, using LMMs for captioning suffers from limitations in understanding structural features, resulting in not enough-detailed annotations or even the hallucination of false information, adversely affecting the accurate multimodal alignment. We propose instead to focus on fewer but useful and detail-rich texts. To this end, we use LMMs not to generate captions for each view, but to generate a small amount of detailed texts that describe hypothetical objects from a given category, covering its subsets and  different instances for structural features. In particular, we leverage GPT\nobreakdash-4~\cite{gpt4} to generate a set of $L$ texts per category $c$ with following prompt:
\begin{quote}
In the context of a 3D objects dataset, generate \textbf{L} text descriptions for the category "\textbf{c}" \textbf{<optional if metadata present:} containing "\textbf{METADATA}"\textbf{>}. Describe one object per text, focusing on its relevant structural features and details. The list should cover a high variety of settings and types for each feature. Do not mention texture, materials, or color.
\end{quote}
Where METADATA, if available, contains possible subsets or information on objects in the given category. With the above prompt, we generate $L$ texts per category in our training set and compute their text embeddings $\{e_{text}^{c,l}\}_{l=0}^L$ through OpenCLIP text encoder $E_t$.
We refer to the generated texts and their embedding, respectively, as text landmarks and landmarks, as they will act as such in building 3D descriptors. Specifically, given a 3D object $x_{3D}^i$ from category $c$ and its views $\{v^{i,r}\}_{r=0}^R$, we build descriptors $d^{i,r}$ with dimension $L$ for each view $r$ by measuring its cosine similarity with respect to all of the landmarks for that category: 
\begin{equation}
    d^{i,r}[l] = e_{image}^{i,r} \cdot e_{text}^{c,l}, l \in 1 ... L
\end{equation}
Each element of the descriptor is then a value between $-1$ and $1$ measuring the similarity of the view with respect to one of the landmarks. When $L<F$, this process builds descriptors that live in a sparser and lower-dimensional per-category space, using the generated texts as landmarks (and thus the name) in the latent space of CLIP. Given two 3D objects $(x_{3D}^1,x_{3D}^2)$ and their descriptors $(d^1,d^2)$, each with shape $(R,L)$, we can now compute $(I2L)^2$ shape similarity leveraging the mean Euclidean distance between descriptors of corresponding views:
\begin{equation}
    \operatorname{3D{\text -}Sim_{(I2L)^2}(x_{3D}^1,x_{3D}^2)} = g\left({\frac{1}{V} \sum_{r=1}^V \sqrt{\sum_{l=1}^L{({d^{1,r}[l] - d^{2,r}[l])^2}}}} \right),\quad g(x) = \frac{1}{1+x}
\end{equation}
where $\operatorname{g(x)}$ maps distances to similarities values between 0 and 1. Despite the increased complexity in its construction, this descriptor offers several advantages. It enables the utilization of CLIP language knowledge alongside visual knowledge, and it does this efficiently by working with fewer, yet more detailed and meaningful texts. This approach also shifts the problem to a lower-dimensional representation space controlled by text landmarks, mitigating biases related to color, texture, and material. As we show in \Cref{fig:3Dto3DRetrieval}, the $(I2L)^2$ similarity succeeds when the image-based similarity falls short, correctly preferring structurally similar samples over samples with similar texture/color.

Lastly, in \Cref{fig:3Dto3DRetrieval} we show that averaging $I2I$ and $(I2L)^2$ similarities is beneficial, allowing us to benefit from CLIP visual features of the $I2I$ method while also mitigating image biases and enforcing structural coherence between similar samples thanks to $(I2L)^2$.

\subsection{Contrastive training with Hard Negative Sampling}
Inspired by, successful methods with 2D and text-2D data~\cite{contrastivelearning,filterdisthard}, we apply an importance sampling approach to increase the weight of harder samples in a batch. We leverage the similarity functions defined in the previous section to define a reweighted contrastive loss to align 2D and 3D data. We define this loss as:
\begin{equation}
\label{eq:loss2D3Dhn}
\small
\operatorname{\mathcal{L}_{(2D,3D){\text -}HN}} = - \frac{1}{2N}  \sum\limits_{i=1}^{N} \log{\frac{\exp{\left(\nicefrac{e_{image}^i \cdot e_{shape}^i}{\tau}\right)}} {\sum\limits_{s=1}^{N}w^{i,s}\exp{\left(\nicefrac{e_{image}^i \cdot e_{shape}^s}{\tau}\right)}}} - \frac{1}{2N}  \sum\limits_{s=1}^{N} \log{\frac{\exp{\left(\nicefrac{e_{image}^s \cdot e_{shape}^s}{\tau}\right)}} {\sum\limits_{i=1}^{N}w^{s,i}\exp{\left(\nicefrac{e_{image}^i \cdot e_{shape}^s}{\tau}\right)}}},
\end{equation}
where weights $w$ for a given batch are defined as:
\begin{equation}
\label{eq:weights}
    w^{i,s} = \frac{(N-1)sim(x_{3D}^i,x_{3D}^s)}{\sum\limits_{k\neq i}sim(x_{3D}^i,x_{3D}^k)},\quad w^{s,i} = \frac{(N-1)sim(x_{3D}^i,x_{3D}^s)}{\sum\limits_{k\neq s}sim(x_{3D}^k,x_{3D}^s)},
\end{equation}
and $sim(\cdot,\cdot)$ is any positive similarity function between 3D samples. For each sample in a batch, the weights consider the mean similarity of all its negative samples. Typically, for each sample, easy negatives significantly outnumber hard negatives within a batch, causing the mean similarity to align with easy negative values. Consequently, easy negatives receive weights close to 1, while harder negatives—characterized by notably higher similarity—are upweighted in the loss function.

As similarity $sim(\cdot,\cdot)$ in \Cref{eq:weights} we can employ both $3D{\text -}Sim_{I2I}$ and $3D{\text -}Sim_{(I2L)^2}$. To enhance training efficiency, we precompute and store similarity values. Notably, we defined the $(I2L)^2$ similarity exclusively for samples within the same category. This choice leverages the observation that hard negatives often belong to the same category, sharing both concept and appearance. This observation allows us to significantly reduce the computational cost and memory requirements for similarity precomputation from $D^2$ (where $D$ is dataset size), to $\sum_{c \in C} |c|^2$ with $|c|$ representing the number of samples in category $c$. We then define $sim(x,y)$ as $3D{\text -}Sim(x,y)$ when the samples are from the same category, while we set it to a constant $\alpha$ otherwise, as to not obtain zero weights in \Cref{eq:weights}.

\section{Experiments}
\subsection{Model and training details}
We employ the pretrained OpenCLIP model, which we freeze, alongside a trainable point cloud encoder. To avoid architectural biases, we adopt the more efficient PointNeXt configuration from the state-of-the-art ULIP\nobreakdash-2~\cite{ulip2}, thus exclusively validating the novel training method. We also train on the same ShapeNet~\cite{shapenet} subset, comprising 55 categories of 3D objects, each represented by 30 equally spaced views and a point cloud sampled from the mesh surface. We normalize point clouds to the unit sphere and apply random translations, rotations, and noise augmentation. The training process utilizes the loss defined in Equation (2), with the AdamW optimizer and a cosine annealing schedule featuring linear warmup.

We train models using the proposed $3D{\text -}Sim_{I2I}$, $3D{\text -}Sim_{(I2L)^2}$ for hard negatives. Additionally, observing the benefits of combining the two similarities, we also train a third model denoted as $3D{\text -}Sim_{\operatorname{AVG}}$. However, instead of directly averaging the similarities, we compute weights for both $3D{\text -}Sim_{I2I}$, $3D{\text -}Sim_{(I2L)^2}$ and average them, as this is more robust to differences in similarity scale between the two employed similarities.
For all experiments, we set $\alpha = 0.25$ and $L=128$, and we ablate the latter in \Cref{sec:ablation}.
\subsection{Experimental results}
We adopt the evaluation pipeline of previous works~\cite{pointclip,pointclipv2,ulip,ulip2} by testing our models on zero-shot 3D classification and standard 3D classification on both ModelNet40~\cite{modelnet40} and the hardest set from ScanObjectNN~\cite{scanobjectnn} (PB\_T50\_RS). Moreover, to evaluate in a more fine-grained manner the learned multimodal space, we also introduce experiments on cross-modal retrieval. For ULIP\nobreakdash-2, we report only available results as the checkpoints for ShapeNet-pretrained models have not been released.

\textbf{Zero-Shot 3D Classification:} Thanks to the multimodal 
alignment, zero-shot 3D classification can directly be performed by computing text embeddings for the sentences "a point cloud of a <c>" for each category $c$, then predicting the one with the maximum similarity with respect to the point cloud embedding of 
\begin{table}[t]
\scriptsize
    \centering
    \begin{tabular}{l l |c c c c|c c}
        ~ & ~ & \multicolumn{4}{c|}{\textbf{Zero-Shot 3D Classification}} & \multicolumn{2}{c}{\textbf{Standard 3D Classification}}\\ 
        ~ & ~ & \multicolumn{2}{c}{ModelNet40} & \multicolumn{2}{c|}{ScanObjectNN} & ModelNet40 & ScanObjectNN \\ \hline
        Model & \#P (M) & top\nobreakdash-1 & top\nobreakdash-5 & top\nobreakdash-1 & top\nobreakdash-5 & top\nobreakdash-1 & top\nobreakdash-1 \\ \hline
        PointCLIP & 0 & 20.2 & 43.7 & 15.4 & 33.9 & 92.0 & 89.1 \\
        CLIP2Point & 86 & 49.4 & 66.5 & 23.32 & 57.5 & 94.2 & 90.6 \\ 
        ULIP-\scriptsize{PointNet++} & 1.5 & 58.4 & 78.2 & 49.9 & 78.8 & 93.5 & 89.7 \\ 
        ULIP-\scriptsize{PointBERT} &  22.1 & 60.4  & 84.0 & 48.5 & 79.9 & 94.1 & 86.4 \\ 
        ReCon & 43.6 & 61.7 & 83.6 & 30.5 & 63.2 & \textbf{94.7} & \textbf{91.3} \\ 
        ULIP 2-\scriptsize{PointNeXt} & 1.4 & 64.5 & 81.3 & - & - & - & 89.5 \\ 
        ULIP 2-\scriptsize{PointBERT} & 22.1 & \textbf{66.4} & \textbf{87.7} & - & - & - & 86.7 \\ \hline
        Ours-\scriptsize{PointNeXt-$I2I$} & 1.4 & 64.2 & \underline{87.1} & 53.1 & 82.4 & 93.2 & 90.1 \\ 
        Ours-\scriptsize{PointNeXt-$(I2L)^2$} & 1.4 & 63.7 & 86.9 & 54.8 & \underline{\textbf{83.9}} & \underline{94.3} & 88.9 \\
        Ours-\scriptsize{PointNeXt-$\operatorname{AVG}$} & 1.4 & \underline{65.9} & 87.0 & \underline{\textbf{55.7}} & 82.6 & 93.7 & \underline{90.2} \\
    \end{tabular}
    \caption{3D classification results. Our best model is underlined, the best overall is in bold.}
    \label{tab:zeroshot}
    \vspace{-1mm}
\end{table}
the considered sample. We report top\nobreakdash-1 accuracy and top\nobreakdash-5 accuracy results in \Cref{tab:zeroshot}. On ModelNet40, our models outperform most previous models despite the lower number of parameters, being on par or better with corresponding ULIP\nobreakdash-2 PointNeXt architectures. Our $\operatorname{AVG}$ model surpasses ReCon, which has $30x$ more parameters and is also competitive with the best ULIP\nobreakdash-2 PointBERT architecture, which has $15x$ more parameters. On ScanObjectNN, our model outperforms all baselines and, based on ULIP\nobreakdash-2 ModelNet improvements with respect to ULIP, we expect it to be comparable to ULIP 2 models too. Note that our models do not train on directly aligning 3D and text data, and thus the performance of our model also shows that, as suggested in Imagebind~\cite{imagebind}, the alignment of all couples of modalities is not needed, as aligning two modalities suffices to indirectly align the others.

\textbf{Standard 3D Classification:} We adopt the community evaluation protocol on 3D classification by concatenating a linear head to the point cloud encoder and finetuning it on classification. Results, reported in \Cref{tab:zeroshot} show that our models obtain competitive performances with respect to state-of-the art models and similar to analogous PointNeXt architectures. This shows that our models learn a meaningful space which serves as a good initialization for fine-tuning on other 3D datasets, obtaining competitive results even with a much lower number of parameters.
\begin{table}[h]
\scriptsize
    \centering
    \begin{tabular}{l l| cc|cc | cc|cc }
        ~ & ~ &\multicolumn{4}{c|}{\textbf{No Background}} & \multicolumn{4}{c}{\textbf{Background}}\\ 
        ~ & ~ & \multicolumn{2}{c|}{Image-to-Shape} & \multicolumn{2}{c|}{Shape-to-Image} & \multicolumn{2}{c}{Image-to-Shape} & \multicolumn{2}{c}{Shape-to-Image} \\ \hline
        Model & \#P (M) & top\nobreakdash-1 & top\nobreakdash-5 & top\nobreakdash-1 & top\nobreakdash-5 & top\nobreakdash-1 & top\nobreakdash-5 & top\nobreakdash-1 & top\nobreakdash-5\\ \hline
        PointCLIP & 0 & 13.6 & 47.3 & 10.1 & 42.5 & 9.9 & 41.2 & 7.4 & 34.8  \\
        CLIP2Point & 86 & 20.4 & 63.2 & 16.5 & 53.7 & 15.0 & 50.1 & 10.9 & 45.2 \\ 
        ULIP-\scriptsize{PointNet++} & 1.5  & 23.9 & 64.6 & 17.2 & 55.9 & 15.1 & 49.2 & 11.4 & 47.3\\ 
        ULIP-\scriptsize{PointBERT} & 22.1 & 25.1 & 65.8 & 17.5 & 55.6 & 14.8 & 51.4 & 12.0 & 47.5\\ 
        ReCon & 43.6 & 31.4 & 72.8 & 22.8 & 63.1 & 20.7 & 56.4 & 19.6 & 53.2\\ \hline
        Ours-\scriptsize{PointNeXt-$I2I$} & 1.4 & 35.2 & 74.4 & 27.7 & 68.1 & \textbf{26.3} & \textbf{60.4} & 22.8 & 56.6\\
        Ours-\scriptsize{PointNeXt-$(I2L)^2$} & 1.4 & 33.3 & 74.4 & 26.8 & 66.8 & 25.3 & 59.9 & 22.2 & 55.5\\
        Ours-\scriptsize{PointNeXt-$\operatorname{AVG}$} & 1.4 & \textbf{35.6} & \textbf{75.1} & \textbf{29.0} & \textbf{68.7} & 25.9 & 60.1 & \textbf{23.3} & \textbf{57.8}\\ 
    \end{tabular}
    \caption{Cross modal retrieval on Pix3D. The best model overall is highlighted in bold.}
    \label{tab:retrieval}
\end{table}

\begin{figure*}[t]
\begin{center}
\includegraphics[width=\linewidth]{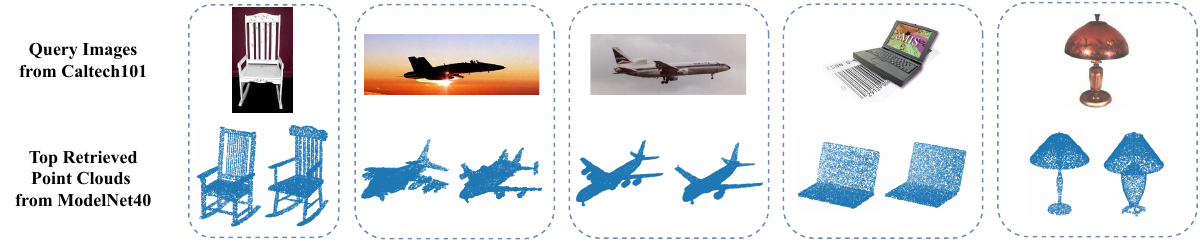}
\end{center}
   \caption{Example of 2D-to-3D cross modal retrieval using our model. Images are from Caltech101, point clouds are from ModelNet40 and are ranked in order of cosine similarity.}
\label{fig:2Dto3DRetrieval}
\end{figure*}
\textbf{Cross-modal retrieval}
While in ULIP~\cite{ulip} authors visually experiment cross-model retrieval with few images from Caltech101 dataset, which for fairness we also test and report in \Cref{fig:2Dto3DRetrieval}, a quantitative analysis is not presented. Unlike 3D classification, which focuses solely on category prediction, cross-modal retrieval assesses the model’s ability to align concepts and fine-grained features across different modalities. To evaluate this aspect, we conduct zero-shot 2D-to-3D and 3D-to-2D retrieval experiments on the Pix3D dataset—a real-world 2D-3D dataset featuring common objects. Notably, Pix3D includes masks for background removal, and we evaluate both the simpler no-background and the more complex background versions. As for classification, we report top\nobreakdash-1 and top\nobreakdash-5 accuracy in \Cref{tab:retrieval}. Our results demonstrate that our models consistently outperform previous approaches in both 2D-to-3D and 3D-to-2D retrieval tasks. The $\operatorname{AVG}$ model excels across most scenarios, except for image-to-shape retrieval with background, where our $I2I$ model exhibits superior performance. Specifically, we achieve an average (top\nobreakdash-1,top\nobreakdash-5) improvement of $(5.2\%,3.95\%)$ on test without backgrounds and $(4.65\%,4.3\%)$ tests with backgrounds. These findings highlight the effectiveness of our approach, showing that the multimodal alignment benefits from our hard negative strategy also at a finer-grained level, leading to a better ability in discriminating correct shapes from images and vice versa.
\section{Ablation Study}
\label{sec:ablation}
\begin{wraptable}{r}{0pt}
\small
$ \begin{array}{c|ccc}
     L & \makecell{Zero \\ Shot} & \makecell{Fine \\ Tuned} & Retrieval\\ \hline
     32 & 52.5 & 93.1 & 16.4 \\
     64 & 58.4 & 92.9 & 20.1 \\
     128 & 63.7 & 94.0 & 23.8 \\
     256 & 63.9 & 93.7 & 24.1 \\
     512 & 63.9 & 94.1 & 23.9 \\
\end{array} $
\caption{Ablation on L value.}
\label{tab:ablation}
\end{wraptable} 
We experiment with different numbers of landmarks per category, and we report top\nobreakdash-1 accuracy results in \Cref{tab:ablation}, on ModelNet40 for zero-shot and finetuned 3D classification and on Pix3D with background for retrieval, averaged over 2D-to-3D and 3D-to-2D, while we report extended results in the supplementary. For this study, we use the $\operatorname{(I2L)^2}$ model to test the effect of landmarks without the contribution of $I2I$ similarity. As we increase the number of generated texts, their diversity decreases and they tend to repetition, becoming unuseful in discriminating different features and structural settings. Consequently, we opt for $L=128$, as it allows a balance between efficiency and performance without compromising accuracy.

\section{Conclusions}
In this study, we presented an teffective method to enhance the 3D alignment to the CLIP model by using CLIP itself, through its visual and textual knowledge. We propose two similarity metrics, $I2I$ and $(I2L)^2$, leveraging CLIP to compare 3D samples and extract 3D hard negatives, and use these to build an effective hard negative weighted contrastive pipeline. With $(I2L)^2$, we also address the challenges of 2D biases (texture, materials, colours) in comparing 3D shapes and we leverage LLM-generated texts in an innovative way, mitigating issues with LMM-generated captions from previous methods. We evaluate our models on zero-shot and finetuned 3D classification, obtaining results comparable or better than previous approaches. Lastly, we evaluate the fine-grained multimodal understanding of our model with cross-modal retrieval, where our models surpass all the previous approaches.
\bibliographystyle{plain}
\bibliography{bib}

\begin{thebibliography}{10}

\bibitem{gpt4}
Josh Achiam, Steven Adler, Sandhini Agarwal, Lama Ahmad, Ilge Akkaya, Florencia~Leoni Aleman, Diogo Almeida, Janko Altenschmidt, Sam Altman, Shyamal Anadkat, et~al.
\newblock Gpt-4 technical report.
\newblock {\em arXiv preprint arXiv:2303.08774}, 2023.

\bibitem{shapenet}
Angel~X Chang, Thomas Funkhouser, Leonidas Guibas, Pat Hanrahan, Qixing Huang, Zimo Li, Silvio Savarese, Manolis Savva, Shuran Song, Hao Su, et~al.
\newblock Shapenet: An information-rich 3d model repository.
\newblock {\em arXiv preprint arXiv:1512.03012}, 2015.

\bibitem{emdcmdproblems}
Ben Fei, Weidong Yang, Wen-Ming Chen, Zhijun Li, Yikang Li, Tao Ma, Xing Hu, and Lipeng Ma.
\newblock Comprehensive review of deep learning-based 3d point cloud completion processing and analysis.
\newblock {\em IEEE Transactions on Intelligent Transportation Systems}, 23(12):22862--22883, 2022.

\bibitem{softclip}
Yuting Gao, Jinfeng Liu, Zihan Xu, Tong Wu, Enwei Zhang, Ke~Li, Jie Yang, Wei Liu, and Xing Sun.
\newblock Softclip: Softer cross-modal alignment makes clip stronger.
\newblock In {\em Proceedings of the AAAI Conference on Artificial Intelligence}, volume~38, pages 1860--1868, 2024.

\bibitem{imagebind}
Rohit Girdhar, Alaaeldin El-Nouby, Zhuang Liu, Mannat Singh, Kalyan~Vasudev Alwala, Armand Joulin, and Ishan Misra.
\newblock Imagebind: One embedding space to bind them all.
\newblock In {\em CVPR}, 2023.

\bibitem{audioclip}
Andrey Guzhov, Federico Raue, J{\"o}rn Hees, and Andreas Dengel.
\newblock Audioclip: Extending clip to image, text and audio.
\newblock In {\em ICASSP 2022-2022 IEEE International Conference on Acoustics, Speech and Signal Processing (ICASSP)}, pages 976--980. IEEE, 2022.

\bibitem{contrastivelearning}
R.~Hadsell, S.~Chopra, and Y.~LeCun.
\newblock Dimensionality reduction by learning an invariant mapping.
\newblock In {\em 2006 IEEE Computer Society Conference on Computer Vision and Pattern Recognition (CVPR'06)}, volume~2, pages 1735--1742, 2006.

\bibitem{clip2point}
Tianyu Huang, Bowen Dong, Yunhan Yang, Xiaoshui Huang, Rynson~W.H. Lau, Wanli Ouyang, and Wangmeng Zuo.
\newblock Clip2point: Transfer clip to point cloud classification with image-depth pre-training.
\newblock In {\em Proceedings of the IEEE/CVF International Conference on Computer Vision (ICCV)}, pages 22157--22167, October 2023.

\bibitem{openclip}
Gabriel Ilharco, Mitchell Wortsman, Ross Wightman, Cade Gordon, Nicholas Carlini, Rohan Taori, Achal Dave, Vaishaal Shankar, Hongseok Namkoong, John Miller, Hannaneh Hajishirzi, Ali Farhadi, and Ludwig Schmidt.
\newblock Openclip, July 2021.
\newblock If you use this software, please cite it as below.

\bibitem{blip2}
Junnan Li, Dongxu Li, Silvio Savarese, and Steven Hoi.
\newblock Blip-2: bootstrapping language-image pre-training with frozen image encoders and large language models.
\newblock In {\em Proceedings of the 40th International Conference on Machine Learning}, ICML'23. JMLR.org, 2023.

\bibitem{llava}
Haotian Liu, Chunyuan Li, Qingyang Wu, and Yong~Jae Lee.
\newblock Visual instruction tuning.
\newblock {\em Advances in neural information processing systems}, 36, 2024.

\bibitem{hardnegrepreslearn1}
Hyun Oh~Song, Yu~Xiang, Stefanie Jegelka, and Silvio Savarese.
\newblock Deep metric learning via lifted structured feature embedding.
\newblock In {\em Proceedings of the IEEE conference on computer vision and pattern recognition}, pages 4004--4012, 2016.

\bibitem{infonce2}
Aaron van~den Oord, Yazhe Li, and Oriol Vinyals.
\newblock Representation learning with contrastive predictive coding.
\newblock {\em arXiv preprint arXiv:1807.03748}, 2018.

\bibitem{recon}
Zekun Qi, Runpei Dong, Guofan Fan, Zheng Ge, Xiangyu Zhang, Kaisheng Ma, and Li~Yi.
\newblock Contrast with reconstruct: Contrastive 3d representation learning guided by generative pretraining.
\newblock In {\em International Conference on Machine Learning (ICML)}, 2023.

\bibitem{filterdisthard}
Filip Radenovic, Abhimanyu Dubey, Abhishek Kadian, Todor Mihaylov, Simon Vandenhende, Yash Patel, Yi~Wen, Vignesh Ramanathan, and Dhruv Mahajan.
\newblock Filtering, distillation, and hard negatives for vision-language pre-training.
\newblock In {\em Proceedings of the IEEE/CVF Conference on Computer Vision and Pattern Recognition (CVPR)}, pages 6967--6977, June 2023.

\bibitem{clip}
Alec Radford, Jong~Wook Kim, Chris Hallacy, Aditya Ramesh, Gabriel Goh, Sandhini Agarwal, Girish Sastry, Amanda Askell, Pamela Mishkin, Jack Clark, et~al.
\newblock Learning transferable visual models from natural language supervision.
\newblock In {\em International conference on machine learning}, pages 8748--8763. PMLR, 2021.

\bibitem{dalle2}
Aditya Ramesh, Prafulla Dhariwal, Alex Nichol, Casey Chu, and Mark Chen.
\newblock Hierarchical text-conditional image generation with clip latents.
\newblock {\em arXiv preprint arXiv:2204.06125}, 1(2):3, 2022.

\bibitem{hardnegsampling}
Robinson, Joshua, Chuang, Ching-Yao, Suvrit Sra, and Stefanie Jegelka.
\newblock Contrastive learning with hard negative samples.
\newblock {\em International Conference on Learning Representations}, 2021.

\bibitem{ic3d}
Cristian Sbrolli, Paolo Cudrano, Matteo Frosi, and Matteo Matteucci.
\newblock Ic3d: Image-conditioned 3d diffusion for shape generation.
\newblock {\em arXiv preprint arXiv:2211.10865}, 2022.

\bibitem{hardnegrepreslearn2}
Florian Schroff, Dmitry Kalenichenko, and James Philbin.
\newblock Facenet: A unified embedding for face recognition and clustering.
\newblock In {\em 2015 IEEE Conference on Computer Vision and Pattern Recognition (CVPR)}, pages 815--823, 2015.

\bibitem{infonce1}
Kihyuk Sohn.
\newblock Improved deep metric learning with multi-class n-pair loss objective.
\newblock In {\em Proceedings of the 30th International Conference on Neural Information Processing Systems}, NIPS'16, page 1857–1865, Red Hook, NY, USA, 2016. Curran Associates Inc.

\bibitem{scanobjectnn}
Mikaela~Angelina Uy, Quang-Hieu Pham, Binh-Son Hua, Duc~Thanh Nguyen, and Sai-Kit Yeung.
\newblock Revisiting point cloud classification: A new benchmark dataset and classification model on real-world data.
\newblock In {\em International Conference on Computer Vision (ICCV)}, 2019.

\bibitem{modelnet40}
Zhirong Wu, Shuran Song, Aditya Khosla, Fisher Yu, Linguang Zhang, Xiaoou Tang, and Jianxiong Xiao.
\newblock 3d shapenets: A deep representation for volumetric shapes.
\newblock In {\em Proceedings of the IEEE conference on computer vision and pattern recognition}, pages 1912--1920, 2015.

\bibitem{videoclip}
Hu~Xu, Gargi Ghosh, Po-Yao Huang, Dmytro Okhonko, Armen Aghajanyan, Florian Metze, Luke Zettlemoyer, and Christoph Feichtenhofer.
\newblock Videoclip: Contrastive pre-training for zero-shot video-text understanding.
\newblock {\em arXiv preprint arXiv:2109.14084}, 2021.

\bibitem{ulip}
Le~Xue, Mingfei Gao, Chen Xing, Roberto Mart{\'\i}n-Mart{\'\i}n, Jiajun Wu, Caiming Xiong, Ran Xu, Juan~Carlos Niebles, and Silvio Savarese.
\newblock Ulip: Learning a unified representation of language, images, and point clouds for 3d understanding.
\newblock In {\em Proceedings of the IEEE/CVF Conference on Computer Vision and Pattern Recognition (CVPR)}, pages 1179--1189, June 2023.

\bibitem{ulip2}
Le~Xue, Ning Yu, Shu Zhang, Junnan Li, Roberto Mart{\'\i}n-Mart{\'\i}n, Jiajun Wu, Caiming Xiong, Ran Xu, Juan~Carlos Niebles, and Silvio Savarese.
\newblock Ulip-2: Towards scalable multimodal pre-training for 3d understanding.
\newblock {\em arXiv preprint arXiv:2305.08275}, 2023.

\bibitem{sigclip}
Xiaohua Zhai, Basil Mustafa, Alexander Kolesnikov, and Lucas Beyer.
\newblock Sigmoid loss for language image pre-training.
\newblock In {\em Proceedings of the IEEE/CVF International Conference on Computer Vision}, pages 11975--11986, 2023.

\bibitem{pointclip}
Renrui Zhang, Ziyu Guo, Wei Zhang, Kunchang Li, Xupeng Miao, Bin Cui, Yu~Qiao, Peng Gao, and Hongsheng Li.
\newblock Pointclip: Point cloud understanding by clip.
\newblock In {\em Proceedings of the IEEE/CVF conference on computer vision and pattern recognition}, pages 8552--8562, 2022.

\bibitem{pointclipv2}
Xiangyang Zhu, Renrui Zhang, Bowei He, Ziyu Guo, Ziyao Zeng, Zipeng Qin, Shanghang Zhang, and Peng Gao.
\newblock Pointclip v2: Prompting clip and gpt for powerful 3d open-world learning.
\newblock In {\em Proceedings of the IEEE/CVF International Conference on Computer Vision}, pages 2639--2650, 2023.

\end{thebibliography}
\end{document}